\documentclass[sigconf]{acmart}
\usepackage{booktabs} 
\usepackage{tabularx} 

\usepackage{listings} 
\usepackage{xcolor} 

\AtBeginDocument{%
  }

\setcopyright{acmlicensed}
\copyrightyear{2025} 
\acmYear{2025} 
\setcopyright{acmlicensed}\acmConference[ICAIF '25]{6th ACM International Conference on AI in Finance}{November 15--18, 2025}{Singapore, Singapore}
\acmBooktitle{6th ACM International Conference on AI in Finance (ICAIF '25), November 15--18, 2025, Singapore, Singapore}
\acmDOI{10.1145/3768292.3770364}

\acmISBN{979-8-4007-2220-2/2025/11}

\begin{document}

\title{FinResearchBench: A Logic Tree based Agent-as-a-Judge Evaluation Framework for Financial Research Agents}


\author{Rui Sun}
\email{sunrui@stepfun.com}
\affiliation{%
  \institution{Stepfun}
  \city{Shanghai}
  \country{China}
}

\author{Zuo Bai}
\email{baizuo@stepfun.com}
\authornote{Corresponding Author}
\affiliation{%
  \institution{Stepfun; FinStep}
  \city{Shanghai}
  \country{China}}

\author{Wentao Zhang}
\email{zhangwentao@stepfun.com}
\affiliation{%
  \institution{Stepfun}
  \city{Shanghai}
  \country{China}}

\author{Yuxiang Zhang}
\email{easonzhang@stepfun.com}
\affiliation{%
  \institution{Stepfun}
  \city{Shanghai}
  \country{China}}

\author{Li Zhao}
\email{zhaoli@stepfun.com}
\affiliation{%
  \institution{Stepfun}
  \city{Shanghai}
  \country{China}}

\author{Shan Sun}
\email{sunshan@finstep.cn}
\affiliation{%
  \institution{FinStep}
  \city{Shanghai}
  \country{China}}

\author{Zhengwen Qiu}
\email{qiuzhengwen@finstep.cn}
\affiliation{%
  \institution{FinStep}
  \city{Shanghai}
  \country{China}}

\renewcommand{\shortauthors}{Sun et al.}

\begin{abstract}
Recently, AI agents are rapidly evolving in intelligence and are widely used in professional research applications, such as STEM, software development, and finance. Among these AI agents, deep research agent is a key category as it can perform long-horizon tasks and solve problems of greater complexity.
However, there are few evaluation frameworks and benchmarks that systematically and automatically investigate the capabilities of these research agents. 
In addition, financial research problems have distinct complexity and subtlety. To fill in the gap, we propose FinResearchBench, which is a logic tree-based Agent-as-a-Judge and targets specifically for the financial research agents. It provides a comprehensive and automatic assessment of the research agents across 7 key types of tasks in the financial research domain.
The contributions of this work are two-folded:
(1) the first and innovative Agent-as-a-Judge system that extracts the logic tree of the research outcome and uses it as the intermediate information to present a comprehensive, reliable, and robust evaluation; (2) finance-oriented that it covers 70 typical financial research questions, spreading across 7 frequently-encountered types of task in the domain.
\end{abstract}



\keywords{AI agents, financial research agents, logic tree, Agent-as-a-Judge}



\maketitle

\section{Introduction}
Large Language Models (LLMs) have been rapidly developing in recent years and demonstrate superior performance in diverse tasks, such as story creation \cite{zhao2023more}, machine translation \cite{10.5555/3692070.3694345}, dialogue system \cite{yi2024survey}, and sentiment analysis \cite{zhang2023sentiment}. Furthermore, LLMs are dramatically evolving in both long text understanding and generation \cite{team2024gemini, Anthropic}. The intrinsic capability of LLM to generate long text is often limited, which is revealed to be bounded by the training data the model has seen during the pretraining and alignment stages \cite{bai2024longwriter}. Yet, long text generation is essential, especially in professional research areas, such as literature, law, STEM, and finance. Numerous methods have been proposed to improve LLM's competence in long text generation, including scaling the generation length by adding long output training data \cite{bai2024longwriter}, performing generation in a multi-agent framework \cite{hong2023metagpt}, and setting up as deep research agents \cite{schmidgall2025agent, zheng2025deepresearcher}.
Deep research agent is a prominent category of AI agents and is getting more and more important, since it could generate a well-articulated, source-attributed research report by autonomously browsing the Internet, calling various tools, collecting information from different sources. The research result is highly valuable and brings a huge improvement in productivity.
Therefore, a comprehensive evaluation of these deep research agents becomes critical. 
Various benchmarks and evaluation methods are proposed, such as HelloBench \cite{que2024hellobench}, Deep Research Bench \cite{bosse2025deep}, and LongEval \cite{wu2025longeval}. However, these methods do not fully utilize the results of the intermediate steps, which could be the deciding factor for the quality of the research outcome. Furthermore, existing benchmarks and evaluation methods are of general purpose and face severe challenges when applied for financial research, since it requires deep understanding of finance-specific terminologies, clauses and domain-specific insights.
In this work, we propose FinResearchBench, a logic tree based Agent-as-a-Judge that is designed specifically for financial research agents. It first extracts the logic tree from the research outcome and then uses the extracted logic tree to build an agentic system to evaluate the financial research agents.
Moreover, FinResearchBench is finance concentrated and covers 7 key types of tasks in the financial research domain.
With the aim of the extracted logic tree as the intermediate step, FinResearchBench is able to provide a comprehensive, reliable and robust evaluation for various financial research agents.

\begin{figure*}[t]
    \centering
    \includegraphics[width=\textwidth]{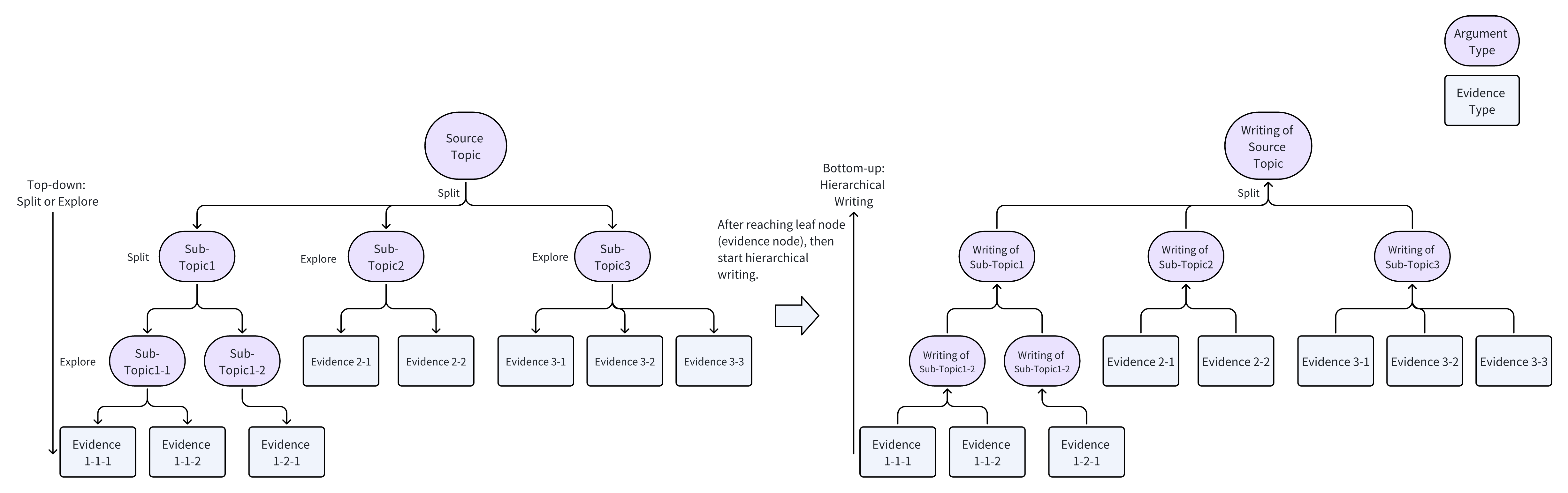}
    \caption{The Illustration of Writing Process}
    \label{fig:writing_process}
\end{figure*}

\section{Related Work}

\subsection{LLM and Deep Research Agent}

Large Language Models (LLMs) have shown remarkable performance in various tasks, such as story creation, machine translation, and dialogue system \cite{zhao2023more, 10.5555/3692070.3694345, yi2024survey, zhang2023sentiment}. In terms of dealing with long text, they also demonstrate superior capability in understanding and processing \cite{team2024gemini, Anthropic}. However, the capability of generating long text progresses rather slowly. Many AI agent systems are then developed including single-agent and multi-agent ones. And it represents a paradigm shift from standalone LLMs towards more autonomous, task-oriented systems \cite{hong2023metagpt, wu2023autogen}.
AI agents largely increase the upper bound of the generation results since they are able to perform more actions, like web browsing, tool calls, and code execution. Additionally, AI agents are capable of processing the obtained information more effectively and accurately, such as Chain-of-Throughts (CoT) method \cite{wei2022chain}, Tree-of-Thoughts (ToT) method \cite{yao2023tree}, 
and ReAct method \cite{yao2023react}. Deep research agent is a growing category of AI agents, since it can orchestrate multiple steps in an autonomous manner, including internet browsing, information retrieval, and tool calling. It is able to provide in-depth research of highly complicated problems, and thus brings huge productivity improvement and high commercial values.

\subsection{General and Domain-Specific Evaluation of Deep Research Agent}
The evaluation of AI agents is critical since it determines if the emergent agents are applicable for real-world problems or not. Moreover, the evaluation will guide the direction for further improvement. Numerous general benchmarks and evaluation methods are proposed, such as AgentBench \cite{liu2023agentbench}, ToolQA \cite{zhuang2023toolqa}, HelloBench \cite{que2024hellobench}, DeepResearch Bench \cite{du2025deepresearch}, and LongEval \cite{wu2025longeval}. These benchmarks are constructed with diverse tools in different environments, enabling extensive evaluation of given agents.
For deep research agent, it is a subcategory of the AI agents and performs multi-step orchestration in an autonomous manner. It produces research report that contains rich information and is well articulated. Many evaluation methods are proposed in the literature, including Agent-as-a-Judge \cite{zhuge2024agent}, and Mind2Web \cite{deng2023mind2web}.

However, these methods are majorly targeted for general domains and inadequate for vertical ones, especially knowledge-heavy domains like law, medicine, STEM and finance. In order to close the gap, many domain-specific benchmarks and evaluation methods are introduced, for instance LegalAgentBench \cite{li2024legalagentbench}, literature review evaluation \cite{tang2024llms}, FinEval \cite{zhang2023fineval} and xbench \cite{chen2025xbench}. These domain-specific benchmarks and evaluations are designed to offer valuable and domain-specific insights for corresponding research.

\section{FinResearchBench}

\subsection{The Design of FinResearchBench: an Agentic Evaluation System}
Financial research is in essence a long text generation problem. And it is impractical to count on human evaluation for long text generation results due to two pitfalls, time-consuming and labor-intensive. Many LLM-as-a-Judge methods were proposed in the literature\cite{zheng2023judging, li2024generation}. However, it remains questionable if directly applying LLM as a judge can robustly and reliably evaluate the results of long text generation, or more specifically in this paper, financial research agents.

The results of financial research agents are open-ended, meaning that there are no reference documents to guide the evaluation, not to mention ground truth to do a simple text-level verification. Asking experienced human experts to give a direct overall quality score is challenging enough. It seems impossible to ask a LLM, no matter how well it is prompted and aligned with humans, to give a direct score and hope it to be accurate and reliable.
HelloBench\cite{que2024hellobench} provides category-specific checklists and asks LLMs to answer yes/no questions for each question on the given checklist. Additionally, financial research problems have distinct intricacy and subtlety, such as financial terminology, logic, cognition and deep insights. More importantly, financial research problems are widely encountered and have huge potential business impacts. Due to the strong correlation between the actual performance of financial research agents and real-world values, we are committed to providing a finance-dedicated evaluation framework featuring hierarchical structures that encompass correctness, informativeness, source attribution, professionlism, among other criteria. And the evaluation framework significantly surpasses existing methods.

\subsubsection{Logic Tree}
In order to draft a comprehensive and insightful professional research document in finance or any other domain, it is mandatory to do a thorough literature review, prepare ideas, and collect supporting evidence before starting to write the full report. Cognitive and linguistic theory defines different stages, i.e. Pre-Writing, Writing and Re-Writing. It defines the planning stage that contains all the preparation steps \cite{rohman1965pre}.
It lays the foundation for a potentially coherent, insightful generation outcome. In this paper, we define the underlying structure of ideas, arguments, and evidences as the logic tree.

It is natural to derive that discovering the underlying logic tree of any financial research outcome and using it as the intermediate step will reduce the difficulty of the evaluation.
Furthermore, the logic tree has some quantifiable metrics, such as depth, width, and number of nodes. Consequently, rule-based metrics can be formulated, such as analysis width, analysis depth, information density, as listed in Table~\ref{tab:model_evaluation_combined}.

    

\subsubsection{Writing Method}
It remains unclear how to extract the underlying logic tree accurately. In this section, we will discuss how professionals write in-depth research. And then it is natural to reverse the writing process and extract an accurate logic tree from a given research report.

Inspired by the cognitive writing theory \cite{rohman1965pre}, there exists different stages like pre-writing and writing. The stage of pre-writing includes multiple operations, such as developing ideas, plans, designs; active enlistment in the cause of an idea; and consecutive logical thinking. And the stage of writing combines the ideas, evidence and words into a fresh, original, well-articulated structure. We show the process in Figure~\ref{fig:writing_process} and formulate it as follows:

\begin{enumerate}
    \item \textbf{Split or Explore}: Analyze the given \textbf{\$Topic\$} and decide to choose \textbf{Split} or \textbf{Explore} in a top-down manner.
    \begin{itemize}
        \item {Split}: if the current \textbf{\$Topic\$} is too broad or should be analyzed from different perspectives, then go with this choice. The split list of sub-topics should comply with the following requirements: (1) the sub-topics should be independent and fully cover all aspects of the source topic; (2) each sub-topic should be specific enough for further information retrieval; (3) the sub-topics should be logically related.
        
        \item {Explore}: if the current \textbf{\$Topic\$} is specific enough, then go with this choice to collect supporting information. A search agent will be followed up to take in the current \textbf{\$Topic\$} and to retrieve relevant information. Subsequently, a well-written summary will be produced based on the \textbf{\$Topic\$} and the relevant information.
    \end{itemize}
    
    \item \textbf{Hierarchical Writing}: in the order of bottom-up, perform writing for each node hierarchically. During the process, writing will be initiated on evidence nodes first, then sub-nodes, and finally the root node. The writing of a certain node will be triggered if it has no sub-node or its sub-nodes are all in ready status. This process yields a distinct argument for each topic node. Consequently, the terms topic and argument are treated as synonymous throughout this paper.
\end{enumerate}

\subsubsection{Logic Tree Extraction}
By reversing the writing method, it is straightforward to propose the extraction method of the logic tree.

\begin{enumerate}
    \item {\textbf{Step1}}: the node of the logic tree has two types: argument type and evidence type, as illustrated Figure~\ref{fig:writing_process}. 
    \begin{itemize}
        \item {Argument Type}: nodes of argument type represent claims, hypotheses, or analytical insights derived from reasoning.
        \item {Evidence Type}: nodes of evidence type represent concrete, factual statements, such as data points, statistical evidence or objectively verifiable facts.
    \end{itemize}

    Leverage a well-prompted LLM to extract the research report into a logic tree, with nodes being classified into argument or evidence type and organized in a hierarchical manner. The extraction must obey the rules to preserve the original report without adding or interpreting any further information.
    
    \item{\textbf{Step2}}: Leverage another well-prompted LLM to evaluate if the argument node can be supported by its subordinate argument nodes or evidence nodes.
\end{enumerate}


\begin{figure*}[t]
    \centering
    \includegraphics[width=0.6\textwidth]{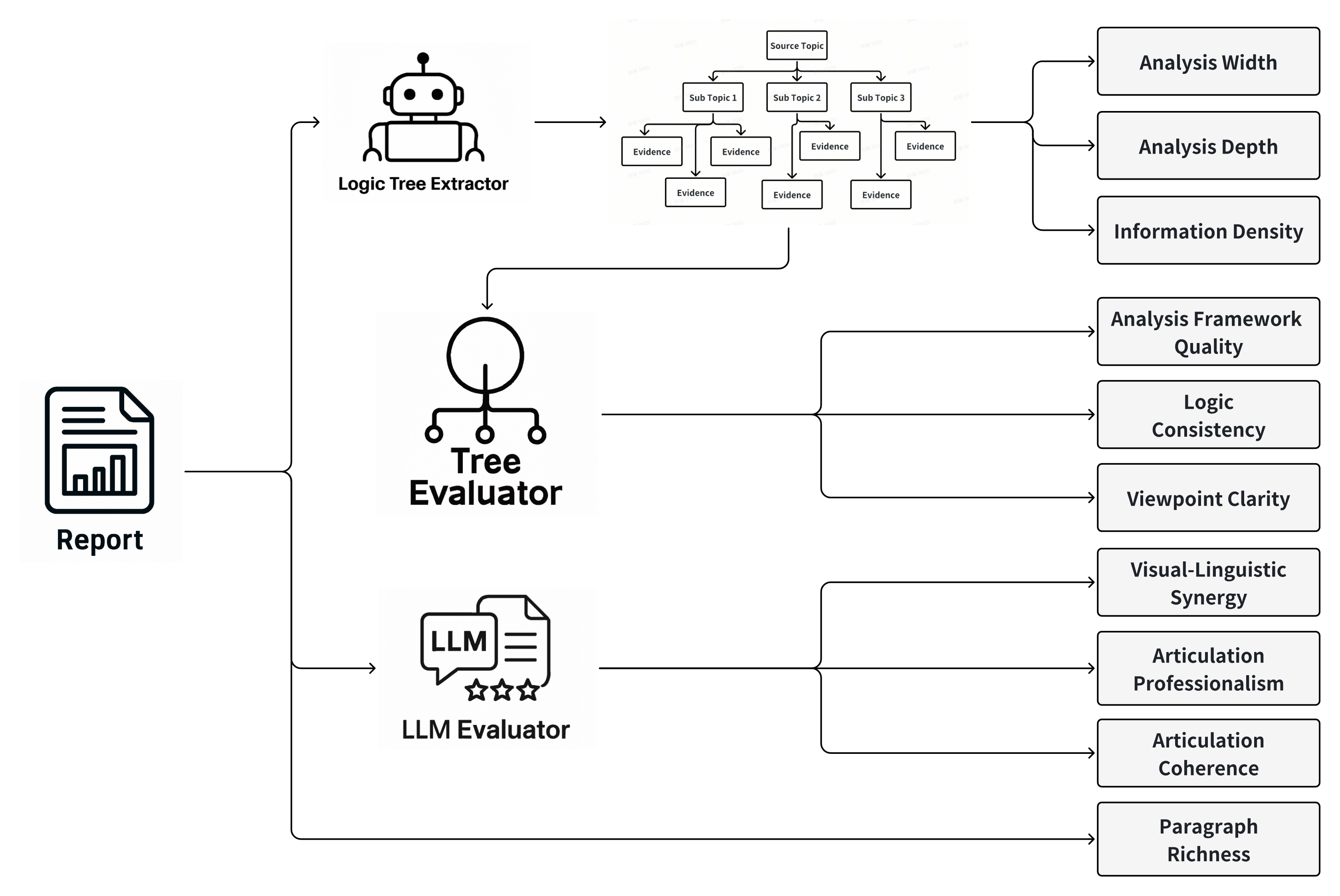}
    \caption{The Illustration of Logic Tree Based Agent-as-a-Judge}
    \label{fig:evaluation_process}
\end{figure*}

\subsubsection{Logic Tree Based Agent-as-a-Judge}
Leveraging the extracted logic tree as the intermediate step, FinResearchBench is developed.
It is an agentic system and essentially a logic tree based Agent-as-a-Judge, as illustrated in Figure~\ref{fig:evaluation_process}.
With the rich intermediate information throughout the writing process, FinResearchBench is able to provide comprehensive, reliable and robust evaluation in an automatic way. As indicated in Table~\ref{tab:model_evaluation_combined}, there are several metrics that are rule derived, either based on the extracted logic tree or not. We explain the rule-based metrics as follows. Notations used in the following section are detailed in Appendix~\ref{app:notations}. 
We note that the hyper-parameters in these scoring functions (e.g., constants and weights) have been determined empirically. This approach is taken to strike a balance between the various analytical dimensions, such as width and depth, while ensuring the resulting scores are sufficiently discriminative.

\begin{enumerate}
    \item \textbf{Analysis Width}: The score of analysis width, $S_{width}$, is calculated based on the average number of children nodes ($\bar{N}_{child}$).
    \begin{equation}
        S_{width} = \min\left(100, \max\left(0, 33.33 \cdot (\bar{N}_{child} - 1)\right)\right)
    \end{equation}

    \item \textbf{Analysis Depth}: The score of analysis depth, $S_{depth}$, is a weighted average of the score from the tree's maximum depth ($S_{D_{max}}$) and the score from the average depth of leaf nodes ($S_{\bar{D}_{leaf}}$).
    \begin{equation}
    \begin{aligned}
        S_{D_{max}} &= \min\left(100, \max\left(0, 25 \cdot (D_{max} - 2)\right)\right) \\
        S_{\bar{D}_{leaf}} &= \min\left(100, \max\left(0, 40 \cdot (\bar{D}_{leaf} - 1.5)\right)\right) \\
        S_{depth} &= 0.4 \cdot S_{D_{max}} + 0.6 \cdot S_{\bar{D}_{leaf}}
    \end{aligned}
    \end{equation}

    \item \textbf{Information Density}: The score of information density, $S_{info}$, is a weighted average of the score from the total number of nodes ($S_{nodes}$) and the score from the evidence density ($S_{density}$).
    \begin{equation}
    \begin{aligned}
        S_{nodes} &= \min\left(100, \max\left(0, 2 \cdot (N_{total} - 5)\right)\right) \\
        R_{evidence} &= \begin{cases}
            \frac{N_{evidence}}{N_{total}} & \text{if } N_{total} > 0 \\
            0 & \text{otherwise}
        \end{cases} \\
        S_{density} &= \min\left(100, 100 \cdot R_{evidence}\right) \\
        S_{info} &= 0.7 \cdot S_{nodes} + 0.3 \cdot S_{density}
    \end{aligned}
    \end{equation}

    \item \textbf{Paragraph Richness}: The score, $S_{rich}$, is a piecewise function of $w$, the average number of words per subtitle.
    \begin{equation}
        S_{rich}(w) = \begin{cases}
            0 & w \leq 0 \\
            0.6w & 0 < w < 100 \\
            60 + 0.08w & 200 \leq w < 500 \\
            100 & 500 \leq w \leq 1000 \\
            \max(60, 100 - 0.05(w - 1000)) & w > 1000
        \end{cases}
    \end{equation}

    \item \textbf{Other Metrics}: For dimensions that are qualitative in nature, such as \textit{Articulation Professionalism}, \textit{Logic Consistency}, and \textit{Viewpoint Clarity}, we employ a Large Language Model (LLM) for evaluation. The LLM is prompted with a detailed rubric to assess and score each of these subjective attributes.
\end{enumerate}





\subsection{Data Construction}
\label{data_construction}
Financial research is skewed and the highly-encountered issues are closely related to monetary returns, such as financial plate investigation and stock-specific analysis. Therefore, there is a huge potential for productivity improvement and real-world values for an accurate, comprehensive and reliable evaluation of these agents.
Additionally, financial research agents have some unique intricacies, such as domain-specific terms, clauses and in-depth knowledge. Therefore, it is crucial to collect sufficiently complex, yet practically applicable financial research tasks.

\begin{itemize}
\item{\textbf{Query Sampling}}: Randomly sample user queries from our online product: \textbf{AI Cashcat} (\url{https://www.cashcat.cn}). A strict privacy protection process is enforced to all raw queries before subsequent steps, including anonymization, personal identifiable information removal and session information disposal.

\item{\textbf{Query Classification}}: We use a general-purpose large language model at hand (\textbf{step-2 model}  from \url{https://www.stepfun.com})  to classify the queries into different categories. In order to determine the category list, we consult 3 domain experts, remove semantically duplicate answers and compile their responses into a union , which is:
\textbf{Stock-Specific Analysis, Event Analysis, Stock Selection from Given Plate, Sector Analysis,
Investment Morning Brief, Post-Market Recap, Expert Opinion}, Stock-Specific Performance, Stock-Specific Movements, Market Performance and Analysis, Plate Performance and Analysis,  Criteria-based Stock Filtering,  Trade Recommendation,  Trending News, Investment Education and Others. 
 The top 7 categories in bold font, out of the total 16 ones, occupy over 95\% online traffic. Therefore, we put the focus on these 7 types of tasks.

\item{\textbf{Data Curation}}: Conduct interviews with domain experts from finance news agencies, research department of security companies, professional investors, and generate 70 questions, 10 typical research problems per category.

\end{itemize}


\section{Experiments and Analysis}

\subsection{Experiment Settings}

This section outlines the experimental setup for our evaluation.

\subsubsection{Core Model}
We utilized Claude 3.7 Sonnet as the foundational Large Language Model (LLM) for the agentic system within the proposed FinResearchBench.

\subsubsection{Agents and Experiments}

We conduct a two-stage evaluation. In the first stage, we run the FinResearchBench on a range of leading international and domestic AI agents as well as our proprietary product, Cashcat Pro (~\url{https://cashcat.cn/?mode=DeepResearch}). \textbf{Cashcat Pro} is a deep research agent developed by FinStep, targeted specifically for the finance sector. It integrates a multi-agent
framework for collaborative analysis, a suite of professional tools for real-time financial data, and cutting-edge AI models to deliver deep, actionable insights and automate the entire research process. In addition, Gemini is powered by Gemini Pro 2.5, while OpenAI DeepResearch is powered by GPT-4o.


In the second stage, we perform a comparison between the top 3 agents versus human expert written report. We collect 20 financial research reports from reputable institutions, focusing on the types of stock-specific analysis and sector analysis. Subsequently, we extract the research topic from each of these reports and ask the top 3 agents to generate corresponding report. Finally, these research outcomes from the agents are compared with the expert written ones, under the same evaluation framework, FinResearchBench, proposed in this paper.

In this way, we obtain a holistic understanding of deep research agents in the financial domain, with not only the performance of different research agents being studied, but also the agents capability versus human experts.

\subsection{Main Results}
The following section presents a comprehensive evaluation of multiple deep research agents, together with expert written reports. Detailed scores are presented in Table~\ref{tab:model_evaluation_combined} and Table~\ref{tab:evaluation_scores_detailed_v3}.

\begin{table*}[h!]
\centering

\caption{Overall Evaluation Results on FinResearchBench}
\label{tab:model_evaluation_combined}

\scriptsize
\setlength{\tabcolsep}{3.0pt}

\begin{tabular}{@{}lcccccccccccc@{}}
\toprule
\textbf{Dimension} & \textbf{Final Score} & \textbf{\begin{tabular}[c]{@{}c@{}}Visual-Linguistic\\ Synergy\end{tabular}} & \textbf{\begin{tabular}[c]{@{}c@{}}Articulation\\ Professionalism\end{tabular}} & \textbf{\begin{tabular}[c]{@{}c@{}}Articulation\\ Coherence\end{tabular}} & \textbf{\begin{tabular}[c]{@{}c@{}}Analysis\\ Framework Quality\end{tabular}} & \textbf{\begin{tabular}[c]{@{}c@{}}Analysis\\ Width\end{tabular}} & \textbf{\begin{tabular}[c]{@{}c@{}}Analysis\\ Depth\end{tabular}} & \textbf{\begin{tabular}[c]{@{}c@{}}Information\\ Density\end{tabular}} & \textbf{\begin{tabular}[c]{@{}c@{}}Logic\\ Consistency\end{tabular}} & \textbf{\begin{tabular}[c]{@{}c@{}}Viewpoint\\ Clarity\end{tabular}} & \textbf{\begin{tabular}[c]{@{}c@{}}Paragraph\\ Richness\end{tabular}} \\
\textbf{Weights} & NA & 10\% & 10\% & 10\% & 10\% & 10\% & 10\% & 10\% & 10\% & 10\% & 10\% \\
\textbf{If Rule Derived} & NA & No & No & No & No & Yes & Yes & Yes & No & No & Yes\\
\midrule
\textbf{Gemini}\raisebox{0.3ex}{\scriptsize *} & \textbf{76.60} & \underline{50.81} & \textbf{95.56} & \textbf{96.72} & \textbf{89.41} & 49.33 & \textbf{82.67} & 42.15 & \underline{89.09} & 87.44 & \underline{82.80} \\
\textbf{OpenAI}\raisebox{0.3ex}{\scriptsize *} & \underline{74.15} & 24.15 & \underline{88.35} & 89.41 & 84.89 & \underline{53.24} & 79.08 & \underline{48.01} & \textbf{90.39} & \textbf{93.10} & \textbf{90.77} \\
\textbf{Cashcat Pro} & 70.73 & \textbf{70.05} & 72.22 & 68.94 & \underline{88.56} & 49.49 & 79.06 & 43.54 & 88.56 & 87.74 & 59.10 \\
\textbf{Minimax}\raisebox{0.3ex}{\scriptsize *} & 67.69 & 29.93 & 72.36 & \underline{90.29} & 85.56 & 52.63 & 75.86 & 31.56 & 88.37 & 85.22 & 65.08 \\
\textbf{Reportify} & 67.18 & 36.06 & 70.68 & 50.60 & 86.08 & \textbf{56.81} & \underline{79.26} & \textbf{51.15} & 87.35 & 85.86 & 67.94 \\
\textbf{Perplexity}\raisebox{0.3ex}{\scriptsize *} & 67.05 & 48.31 & 57.68 & 83.03 & 81.29 & 50.95 & 71.93 & 30.79 & 88.93 & 88.79 & 68.77 \\
\textbf{Genspark}\raisebox{0.3ex}{\scriptsize *} & 64.30 & 10.60 & 66.46 & 82.65 & 82.62 & 50.66 & 78.73 & 40.06 & 87.07 & \underline{89.13} & 54.98 \\
\textbf{AutoGLM} & 59.98 & 0 & 49.06 & 56.82 & 78.31 & 49.35 & 77.05 & 38.16 & 88.56 & \underline{89.13} & 73.26 \\
\bottomrule
\end{tabular}
\begin{minipage}{\linewidth}
    \vspace{2pt}
    \hspace{8pt} \textsuperscript{*}Not officially accessible in Chinese Mainland.
\end{minipage}
\end{table*}

\begin{table*}[h!]
\centering
\caption{Evaluation Results against Human Written Reports}
\label{tab:evaluation_scores_detailed_v3}

\scriptsize 
\setlength{\tabcolsep}{2.5pt} 

\begin{tabular}{@{}lcccccccccccc@{}}
\toprule
\textbf{Dimension} & \textbf{Final Score} & \textbf{\begin{tabular}[c]{@{}c@{}}Visual-Linguistic\\ Synergy\end{tabular}} & \textbf{\begin{tabular}[c]{@{}c@{}}Articulation\\ Professionalism\end{tabular}} & \textbf{\begin{tabular}[c]{@{}c@{}}Articulation\\ Coherence\end{tabular}} & \textbf{\begin{tabular}[c]{@{}c@{}}Analysis\\ Framework Quality\end{tabular}} & \textbf{\begin{tabular}[c]{@{}c@{}}Analysis\\ Width\end{tabular}} & \textbf{\begin{tabular}[c]{@{}c@{}}Analysis\\ Depth\end{tabular}} & \textbf{\begin{tabular}[c]{@{}c@{}}Information\\ Density\end{tabular}} & \textbf{\begin{tabular}[c]{@{}c@{}}Logic\\ Consistency\end{tabular}} & \textbf{\begin{tabular}[c]{@{}c@{}}Viewpoint\\ Clarity\end{tabular}} & \textbf{\begin{tabular}[c]{@{}c@{}}Paragraph\\ Richness\end{tabular}} \\
\textbf{Weights} & NA & 10\% & 10\% & 10\% & 10\% & 10\% & 10\% & 10\% & 10\% & 10\% & 10\% \\
\textbf{If Rule Derived} & NA & No & No & No & No & Yes & Yes & Yes & No & No & Yes\\
\midrule
\textbf{Expert Written Report} & \textbf{81.43} & \textbf{100.00} & 77.29 & 75.85 & 85.24 & \textbf{55.77} & \textbf{83.31} & 46.15 & \textbf{90.98} & 99.76 & \textbf{99.96} \\
\textbf{Gemini}\raisebox{0.3ex}{\scriptsize *} & 77.05 & 46.87 & \textbf{88.13} & \textbf{94.93} & 85.80 & 51.00 & 81.46 & 47.66 & 90.64 & 99.80 & 84.25 \\
\textbf{OpenAI}\raisebox{0.3ex}{\scriptsize *} & 72.03 & 22.13 & 76.40 & 82.47 & 83.07 & 50.29 & 79.36 & \textbf{49.82} & 88.58 & \textbf{99.87} & 88.27 \\
\textbf{Cashcat Pro} & 71.86 & 75.00 & 65.27 & 65.47 & \textbf{87.13} & 51.28 & 77.56 & 46.68 & 90.28 & 99.07 & 60.87 \\
\bottomrule
\end{tabular}
\begin{minipage}{\linewidth}
    \vspace{2pt}
    \hspace{4pt} \textsuperscript{*}Not officially accessible in Chinese Mainland.
\end{minipage}
\end{table*}

\subsubsection{Analysis of Overall Performance on FinResearchBench}
Table~\ref{tab:model_evaluation_combined} presents the overall evaluation results of various agents on FinResearchBench. The key findings are as follows:
\begin{itemize}
    \item \textbf{Overall Ranking:} Among all agents,  \textbf{Gemini} achieves the highest \textit{Final Score} (76.60), establishing as the top-performing agent in this benchmark. \textbf{OpenAI} is close behind with the second-highest score (74.14).

    \item \textbf{Gemini's Strengths:} Gemini's leading position is predominantly driven by its outstanding performance in articulation and analysis. It scores highest in \textit{Articulation Professionalism} (95.56), \textit{Articulation Coherence} (96.72), \textit{Analysis Framework Quality} (89.41), and \textit{Analysis Depth} (82.67), indicating its strong capability to generate professional, coherent, well-structured, and in-depth financial reports.
    
    \item \textbf{OpenAI's Strengths:} OpenAI agent excels in different dimensions. It delivers the top scores in \textit{Logic Consistency} (90.39), \textit{Viewpoint Clarity} (93.10), and \textit{Paragraph Richness} (90.77). This suggests that its outputs are highly logical, clear, and contextually rich.

    \item \textbf{Cashcat Pro's Strengths:} It demonstrates highly competitive performance and ranks in the top tier. It exhibits a unique, specialized skill set that differentiates it from other leading agents. Cashcat Pro shows exceptional performance in \textit{Visual-Linguistic Synergy} and excels at producing highly structured and methodologically sound reports. Furthermore, among the agents accessible in Chinese Mainland, it leads by a clear margin.

\end{itemize}

\subsubsection{Comparative Analysis Against Human Experts}
Table~\ref{tab:evaluation_scores_detailed_v3} provides a direct comparison between the top 3 agents and expert written reports. This analysis reveals a more nuanced view of agent capabilities.
\begin{itemize}
    \item \textbf{Human Benchmark:} \textbf{Expert Written Report} is the leading performer with the highest \textit{Final Score} (81.43), confirming that human experts still produce the highest quality.
    
    \item \textbf{Dimensions of Human Superiority:} The expert's dominance lies in dimensions 
    requiring holistic understanding and synthesis. The expert achieves a perfect score in \textit{Visual-Linguistic Synergy} (100.00) and a near-perfect score in \textit{Paragraph Richness} (99.96). Additionally, the expert leads in \textit{Analysis Width} (55.77), \textit{Analysis Depth} (83.31), and \textit{Logic Consistency} (90.98).
    
    \item \textbf{Conclusion:} In summary, the evaluation results indicate that while no single AI agent surpasses human experts in terms of the overall intelligence, different agents exhibit unique strengths. The path forward to develop a superior deep research agent lies not only in improving individual metrics, but also in integrating these disparate strengths to a further extend.
\end{itemize}

\subsection{Effectiveness of the Logic Tree Extraction}
The accuracy of our proposed method, the logic tree based Agent-as-a-Judge, largely depends on the accuracy of the extraction of the logic tree. Thus, we design a corresponding experiment to validate the effectiveness of the logic tree extraction.


\subsubsection{Objective}
As the efficacy and reliability of our report evaluation methodology largely hinges upon the precision of the extracted logic tree, we aim to quantify the accuracy of the logic tree extraction using clearly defined metrics.

\subsubsection{Methodology}
As shown in Figure ~\ref{fig:writing_process}, the research process is composed of 2 steps: top-down \textbf{Split or Explore} strategy, and bottom-up \textbf{Hierarchical Writing}. It is intuitive to run the 2 steps in a consecutive manner and pack the outcomes of the second step and the first step to form a pair of report and ground-truth (GT) logic tree.

For each pair of report-GT logic tree, the logic tree extraction method will take over and extract corresponding logic tree from the report. And then, the similarity between the GT logic tree and the extracted one will be used to quantify the accuracy of the proposed logic tree extraction method.

\subsubsection{Ground Truth Generation}
We use some random queries as described in Section~\ref{data_construction} to generate many pairs of report-GT logic tree. Furthermore, some rules are applied to pick out 100 pairs with the GT logic tree having different numbers of depth and width to enforce diversity on the generated dataset.

\subsubsection{Similarity Metrics}
We introduce 3 metrics to comprehensively compare the GT logic tree and the extracted logic tree:

\begin{itemize}
    \item \textbf{Total Nodes Similarity:} the similarity between the total number of nodes in the extracted logic tree and the GT. Here, \( N_a \) and \( N_b \) denote the total number of nodes in the extracted logic tree and the GT logic tree, respectively.
    \[
    S_{\text{nodes}} = 1 - \frac{|N_a - N_b|}{\max(N_a, N_b)}
    \]
    \item \textbf{Depth Similarity:} the similarity between the average depths of the extracted logic tree and the GT. \( d_a \) and \( d_b \) are the average depths of the extracted and GT logic trees, respectively.
    \[
    S_{\text{depth}} = 1 - \frac{|d_a - d_b|}{\max(d_a, d_b)}
    \]
    \item \textbf{Width Similarity:} the similarity between the average number of children (width) per node in the extracted logic tree and the GT. \( w_a \) and \( w_b \) refer to the average number of children per node in the extracted and GT, respectively.
    \[
    S_{\text{width}} = 1 - \frac{|w_a - w_b|}{\max(w_a, w_b)}
    \]
\end{itemize}

\subsubsection{Results}
The metrics are computed and listed as follows:

\begin{center}
\begin{tabular}{|c|c|}
\hline
\textbf{Dimension} & \textbf{Similarity Score} \\
\hline
Total Nodes Similarity & 0.54 \\
Depth Similarity & 0.85 \\
Width Similarity & 0.76 \\
\hline
\textbf{Average Similarity} & \textbf{0.72} \\
\hline
\end{tabular}
\end{center}

A detailed analysis indicates a significant drop in the extraction accuracy when the GT logic tree is either too large or too small. Therefore, we check the metric scores in different bins in terms of the nodes count of the GT logic tree.

\begin{center}
\begin{tabular}{|c|c|c|c|}
\hline
\textbf{Dimension} & \textbf{[0 - 25)} & \textbf{[25 - 50)} & \textbf{[50, +$\infty)$} \\
\hline
Total Nodes Similarity & 0.37 & 0.78 & 0.48 \\
Depth Similarity & 0.81 & 0.89 & 0.85 \\
Width Similarity & 0.66 & 0.82 & 0.81 \\
\hline
\textbf{Average Similarity} & \textbf{0.61} & \textbf{0.83} & \textbf{0.71} \\
\hline
\end{tabular}
\end{center}

\subsubsection{Insights and Conclusions}
The evaluation clearly demonstrates that the logic tree extraction is robust for moderately complex structures (25-50 nodes), with an average similarity of 0.83. However, the extraction accuracy significantly diminishes for smaller (<25 nodes, similarity 0.61) and larger trees (>50 nodes, similarity 0.71). This underscores the necessity to optimize the extraction step, tailored specifically towards very concise or extensively detailed reports to ensure reliability of the proposed method.

To further analyze the applicability of the logic tree, additional statistics are calculated:

\begin{center}
\begin{tabular}{|c|c|c|c|}
\hline
\textbf{Dimension} & \textbf{[0 - 25)} & \textbf{[25 - 50)} & \textbf{[50, +$\infty)$} \\
\hline
Avg. Extracted Nodes Count & 37 & 49 & 48 \\
Min. Word Count & 403 & 10901 & 31946 \\
Max. Word Count & 9103 & 16532 & 50416 \\
Spearman Correlation & \multicolumn{2}{c|}{0.48 (p-value: 0.0001)} & \\
\hline
\end{tabular}
\end{center}

We perform a \textbf{correlation analysis} between the total number of nodes from the extracted logic tree and the GT. As detailed in the above table, the analysis yields a \textbf{Spearman correlation coefficient} of $r = 0.48$, with a highly statistically significant \textit{p}-value of less than $0.0001$.

This \textbf{moderate positive correlation} indicates that, overall speaking, an increase in the size of GT logic tree is associated with an increase in the size of the extracted logic tree. This finding supports our hypothesis that the logic tree method is sensitive to the underlying complexity of the source report.

More specifically, this relationship is particularly effective for documents up to approximately 16,000 words. As shown in the table, when the document word count increases from the first bin (max 9,103 words) to the second bin (max 16,532 words), the \textit{Avg. Extracted Nodes Count} rises accordingly from 37 to 49. This demonstrates that for short to medium-length documents, our method robustly captures the increasing complexity by producing a larger and more complicated logic tree.

However, the data also suggests a \textbf{potential limitation} for extremely long report. As the word count extends into the third bin (over 30,000 words), the average number of extracted nodes plateaus and slightly decreases to 48. This implies that while the logic tree method serves as an accurate indicator of complexity for a majority of documents, its ability to differentiate the level of complexity diminishes for exceptionally long and elaborate reports.

\section{Conclusions}
\subsection{Main Conclusions}
In this paper, we introduce FinResearchBench, which is the first logic tree-based Agent-as-a-Judge system that is capable of performing an automatic evaluation to financial research agents and providing reliable and robust results. In addition, we collect 70 head research questions, spreading across 7 key types of tasks in the financial research domain. 
The problems in the domain of financial research have big potentials in investment activities, making the proposed FinResearchBench to have huge real-world values.
We expect FinResearchBench to become a useful tool for the developers of any deep research agents in the financial domain and help those agents to proceed even further.

\subsection{Challenges and Future Works}
Despite the promising results, this study identifies several limitations that need more attention in future research:

\textbf{Enhancing Multimodal Generation Capabilities:} A significant gap persists in the visualization capabilities between the current AI agents and human experts. Even though current advanced agents can generate text effectively, their ability to create and integrate meaningful charts, graphs, and diagrams remain nascent. This highlights a critical direction for future work: advancing the multimodal generation capability. 
Future research should focus on enabling agents not only to analyze visual data but also to generate informative and aesthetically pleasing visuals that are deeply synergistic with the textual content.

\textbf{Improving Logic Tree Extraction for Document of Varied Length:} Our analysis indicates that the current logic tree extraction method shows varied efficacy across document of different lengths. 
The extraction method is suboptimal for very short and very long reports. Future work should aim to enhance the robustness of the extraction algorithm, making it less sensitive to the length of the document, and thus providing more reliable intermediate results to further improve the  end-to-end evaluation.

\textbf{Moving from Logic Trees to Logic Graphs:} Furthermore, a more fundamental limitation lies in the choice of a tree structure itself to represent a document's logic. While trees excel at capturing hierarchical, top-down arguments, they inherently struggle to model the complex, network-like nature of sophisticated reasoning. Real-world argumentation is rarely a strict hierarchy; it often involves cross-references where a single piece of evidence supports multiple disparate claims, feedback loops, or premises that jointly support a conclusion in a non-linear fashion.

A logic graph, particularly a directed acyclic graph (DAG), offers a far more robust and complete representational framework. Such a structure can naturally capture these many-to-many relationships, allowing for a more nuanced and accurate evaluation of a document's overall coherence and logical integrity. Therefore, a pivotal direction for future work is to move beyond tree-based models and develop novel methods for extracting and analyzing these more comprehensive logic graphs. This would represent a significant step towards an AI that can truly understand the intricate web of arguments presented in a document.

\begin{acks}
This paper was supported by "2025 Special Program for Supporting Innovative Development in Leading Industries (AI Track) under the High-Quality Industrial Development Initiative" (Project Name: Finstep Finsmart Intelligent Service Platform; Project ID: 2025-GZL-RGZN-01024) from Shanghai Municipal Commission of Economy and Informatization, Shanghai, China.
\end{acks}

\bibliographystyle{ACM-Reference-Format}
\bibliography{sample-base}


\begin{thebibliography}{29}


\ifx \showCODEN    \undefined \def \showCODEN     #1{\unskip}     \fi
\ifx \showISBNx    \undefined \def \showISBNx     #1{\unskip}     \fi
\ifx \showISBNxiii \undefined \def \showISBNxiii  #1{\unskip}     \fi
\ifx \showISSN     \undefined \def \showISSN      #1{\unskip}     \fi
\ifx \showLCCN     \undefined \def \showLCCN      #1{\unskip}     \fi
\ifx \shownote     \undefined \def \shownote      #1{#1}          \fi
\ifx \showarticletitle \undefined \def \showarticletitle #1{#1}   \fi
\ifx \showURL      \undefined \def \showURL       {\relax}        \fi
\providecommand\bibfield[2]{#2}
\providecommand\bibinfo[2]{#2}
\providecommand\natexlab[1]{#1}
\providecommand\showeprint[2][]{arXiv:#2}

\bibitem[Anthropic(2024)]%
        {Anthropic}
\bibfield{author}{\bibinfo{person}{Anthropic}.} \bibinfo{year}{2024}\natexlab{}.
\newblock \bibinfo{title}{Anthropic: Introducing Claude 3.5 Sonnet}.
\newblock
\urldef\tempurl%
\url{https://www.anthropic.com/news/claude-3-5-sonnet}
\showURL{%
\tempurl}


\bibitem[Bai et~al\mbox{.}(2024)]%
        {bai2024longwriter}
\bibfield{author}{\bibinfo{person}{Yushi Bai}, \bibinfo{person}{Jiajie Zhang}, \bibinfo{person}{Xin Lv}, \bibinfo{person}{Linzhi Zheng}, \bibinfo{person}{Siqi Zhu}, \bibinfo{person}{Lei Hou}, \bibinfo{person}{Yuxiao Dong}, \bibinfo{person}{Jie Tang}, {and} \bibinfo{person}{Juanzi Li}.} \bibinfo{year}{2024}\natexlab{}.
\newblock \bibinfo{title}{LongWriter: Unleashing 10,000+ Word Generation from Long Context LLMs}.
\newblock
\showeprint[arxiv]{2408.07055}~[cs.CL]
\urldef\tempurl%
\url{https://arxiv.org/abs/2408.07055}
\showURL{%
\tempurl}


\bibitem[Chen et~al\mbox{.}(2025)]%
        {chen2025xbench}
\bibfield{author}{\bibinfo{person}{Kaiyuan Chen}, \bibinfo{person}{Yixin Ren}, \bibinfo{person}{Yang Liu}, \bibinfo{person}{Xiaobo Hu}, \bibinfo{person}{Haotong Tian}, \bibinfo{person}{Tianbao Xie}, \bibinfo{person}{Fangfu Liu}, \bibinfo{person}{Haoye Zhang}, \bibinfo{person}{Hongzhang Liu}, \bibinfo{person}{Yuan Gong}, \bibinfo{person}{Chen Sun}, \bibinfo{person}{Han Hou}, \bibinfo{person}{Hui Yang}, \bibinfo{person}{James Pan}, \bibinfo{person}{Jianan Lou}, \bibinfo{person}{Jiayi Mao}, \bibinfo{person}{Jizheng Liu}, \bibinfo{person}{Jinpeng Li}, \bibinfo{person}{Kangyi Liu}, \bibinfo{person}{Kenkun Liu}, \bibinfo{person}{Rui Wang}, \bibinfo{person}{Run Li}, \bibinfo{person}{Tong Niu}, \bibinfo{person}{Wenlong Zhang}, \bibinfo{person}{Wenqi Yan}, \bibinfo{person}{Xuanzheng Wang}, \bibinfo{person}{Yuchen Zhang}, \bibinfo{person}{Yi-Hsin Hung}, \bibinfo{person}{Yuan Jiang}, \bibinfo{person}{Zexuan Liu}, \bibinfo{person}{Zihan Yin}, \bibinfo{person}{Zijian Ma}, {and} \bibinfo{person}{Zhiwen Mo}.}
  \bibinfo{year}{2025}\natexlab{}.
\newblock \bibinfo{title}{xbench: Tracking Agents Productivity Scaling with Profession-Aligned Real-World Evaluations}.
\newblock
\showeprint[arxiv]{2506.13651}~[cs.LG]
\urldef\tempurl%
\url{https://arxiv.org/abs/2506.13651}
\showURL{%
\tempurl}


\bibitem[Deng et~al\mbox{.}(2023)]%
        {deng2023mind2web}
\bibfield{author}{\bibinfo{person}{Xiang Deng}, \bibinfo{person}{Yu Gu}, \bibinfo{person}{Boyuan Zheng}, \bibinfo{person}{Shijie Chen}, \bibinfo{person}{Samuel Stevens}, \bibinfo{person}{Boshi Wang}, \bibinfo{person}{Huan Sun}, {and} \bibinfo{person}{Yu Su}.} \bibinfo{year}{2023}\natexlab{}.
\newblock \showarticletitle{MIND2WEB: towards a generalist agent for the web}. In \bibinfo{booktitle}{\emph{Proceedings of the 37th International Conference on Neural Information Processing Systems}} (New Orleans, LA, USA) \emph{(\bibinfo{series}{NIPS '23})}. \bibinfo{publisher}{Curran Associates Inc.}, \bibinfo{address}{Red Hook, NY, USA}, Article \bibinfo{articleno}{1220}, \bibinfo{numpages}{24}~pages.
\newblock


\bibitem[Du et~al\mbox{.}(2025)]%
        {du2025deepresearch}
\bibfield{author}{\bibinfo{person}{Mingxuan Du}, \bibinfo{person}{Benfeng Xu}, \bibinfo{person}{Chiwei Zhu}, \bibinfo{person}{Xiaorui Wang}, {and} \bibinfo{person}{Zhendong Mao}.} \bibinfo{year}{2025}\natexlab{}.
\newblock \bibinfo{title}{DeepResearch Bench: A Comprehensive Benchmark for Deep Research Agents}.
\newblock
\showeprint[arxiv]{2506.11763}~[cs.CL]
\urldef\tempurl%
\url{https://arxiv.org/abs/2506.11763}
\showURL{%
\tempurl}


\bibitem[FutureSearch et~al\mbox{.}(2025)]%
        {bosse2025deep}
\bibfield{author}{\bibinfo{person}{FutureSearch}, \bibinfo{person}{:}, \bibinfo{person}{Nikos~I. Bosse}, \bibinfo{person}{Jon Evans}, \bibinfo{person}{Robert~G. Gambee}, \bibinfo{person}{Daniel Hnyk}, \bibinfo{person}{Peter Mühlbacher}, \bibinfo{person}{Lawrence Phillips}, \bibinfo{person}{Dan Schwarz}, {and} \bibinfo{person}{Jack Wildman}.} \bibinfo{year}{2025}\natexlab{}.
\newblock \bibinfo{title}{Deep Research Bench: Evaluating AI Web Research Agents}.
\newblock
\showeprint[arxiv]{2506.06287}~[cs.AI]
\urldef\tempurl%
\url{https://arxiv.org/abs/2506.06287}
\showURL{%
\tempurl}


\bibitem[Guo et~al\mbox{.}(2024)]%
        {zhang2023fineval}
\bibfield{author}{\bibinfo{person}{Xin Guo}, \bibinfo{person}{Haotian Xia}, \bibinfo{person}{Zhaowei Liu}, \bibinfo{person}{Hanyang Cao}, \bibinfo{person}{Zhi Yang}, \bibinfo{person}{Zhiqiang Liu}, \bibinfo{person}{Sizhe Wang}, \bibinfo{person}{Jinyi Niu}, \bibinfo{person}{Chuqi Wang}, \bibinfo{person}{Yanhui Wang}, \bibinfo{person}{Xiaolong Liang}, \bibinfo{person}{Xiaoming Huang}, \bibinfo{person}{Bing Zhu}, \bibinfo{person}{Zhongyu Wei}, \bibinfo{person}{Yun Chen}, \bibinfo{person}{Weining Shen}, {and} \bibinfo{person}{Liwen Zhang}.} \bibinfo{year}{2024}\natexlab{}.
\newblock \bibinfo{title}{FinEval: A Chinese Financial Domain Knowledge Evaluation Benchmark for Large Language Models}.
\newblock
\showeprint[arxiv]{2308.09975}~[cs.CL]
\urldef\tempurl%
\url{https://arxiv.org/abs/2308.09975}
\showURL{%
\tempurl}


\bibitem[Hong et~al\mbox{.}(2024)]%
        {hong2023metagpt}
\bibfield{author}{\bibinfo{person}{Sirui Hong}, \bibinfo{person}{Mingchen Zhuge}, \bibinfo{person}{Jonathan Chen}, \bibinfo{person}{Xiawu Zheng}, \bibinfo{person}{Yuheng Cheng}, \bibinfo{person}{Jinlin Wang}, \bibinfo{person}{Ceyao Zhang}, \bibinfo{person}{Zili Wang}, \bibinfo{person}{Steven Ka~Shing Yau}, \bibinfo{person}{Zijuan Lin}, \bibinfo{person}{Liyang Zhou}, \bibinfo{person}{Chenyu Ran}, \bibinfo{person}{Lingfeng Xiao}, \bibinfo{person}{Chenglin Wu}, {and} \bibinfo{person}{Jürgen Schmidhuber}.} \bibinfo{year}{2024}\natexlab{}.
\newblock \showarticletitle{MetaGPT: Meta Programming for A Multi-Agent Collaborative Framework}. In \bibinfo{booktitle}{\emph{ICLR}}.
\newblock
\urldef\tempurl%
\url{https://openreview.net/forum?id=VtmBAGCN7o}
\showURL{%
\tempurl}


\bibitem[Li et~al\mbox{.}(2024b)]%
        {li2024generation}
\bibfield{author}{\bibinfo{person}{Dawei Li}, \bibinfo{person}{Bohan Jiang}, \bibinfo{person}{Liangjie Huang}, \bibinfo{person}{Alimohammad Beigi}, \bibinfo{person}{Chengshuai Zhao}, \bibinfo{person}{Zhen Tan}, \bibinfo{person}{Amrita Bhattacharjee}, \bibinfo{person}{Yuxuan Jiang}, \bibinfo{person}{Canyu Chen}, \bibinfo{person}{Tianhao Wu}, {et~al\mbox{.}}} \bibinfo{year}{2024}\natexlab{b}.
\newblock \showarticletitle{From generation to judgment: Opportunities and challenges of llm-as-a-judge}.
\newblock \bibinfo{journal}{\emph{arXiv preprint arXiv:2411.16594}} (\bibinfo{year}{2024}).
\newblock


\bibitem[Li et~al\mbox{.}(2024a)]%
        {li2024legalagentbench}
\bibfield{author}{\bibinfo{person}{Haitao Li}, \bibinfo{person}{Junjie Chen}, \bibinfo{person}{Jingli Yang}, \bibinfo{person}{Qingyao Ai}, \bibinfo{person}{Wei Jia}, \bibinfo{person}{Youfeng Liu}, \bibinfo{person}{Kai Lin}, \bibinfo{person}{Yueyue Wu}, \bibinfo{person}{Guozhi Yuan}, \bibinfo{person}{Yiran Hu}, {et~al\mbox{.}}} \bibinfo{year}{2024}\natexlab{a}.
\newblock \showarticletitle{LegalAgentBench: Evaluating LLM Agents in Legal Domain}.
\newblock \bibinfo{journal}{\emph{arXiv preprint arXiv:2412.17259}} (\bibinfo{year}{2024}).
\newblock


\bibitem[Liu et~al\mbox{.}(2024)]%
        {liu2023agentbench}
\bibfield{author}{\bibinfo{person}{Xiao Liu}, \bibinfo{person}{Hao Yu}, \bibinfo{person}{Hanchen Zhang}, \bibinfo{person}{Yifan Xu}, \bibinfo{person}{Xuanyu Lei}, \bibinfo{person}{Hanyu Lai}, \bibinfo{person}{Yu Gu}, \bibinfo{person}{Hangliang Ding}, \bibinfo{person}{Kaiwen Men}, \bibinfo{person}{Kejuan Yang}, \bibinfo{person}{Shudan Zhang}, \bibinfo{person}{Xiang Deng}, \bibinfo{person}{Aohan Zeng}, \bibinfo{person}{Zhengxiao Du}, \bibinfo{person}{Chenhui Zhang}, \bibinfo{person}{Sheng Shen}, \bibinfo{person}{Tianjun Zhang}, \bibinfo{person}{Yu Su}, \bibinfo{person}{Huan Sun}, \bibinfo{person}{Minlie Huang}, \bibinfo{person}{Yuxiao Dong}, {and} \bibinfo{person}{Jie Tang}.} \bibinfo{year}{2024}\natexlab{}.
\newblock \showarticletitle{AgentBench: Evaluating {LLM}s as Agents}. In \bibinfo{booktitle}{\emph{The Twelfth International Conference on Learning Representations}}.
\newblock
\urldef\tempurl%
\url{https://openreview.net/forum?id=zAdUB0aCTQ}
\showURL{%
\tempurl}


\bibitem[Que et~al\mbox{.}(2024)]%
        {que2024hellobench}
\bibfield{author}{\bibinfo{person}{Haoran Que}, \bibinfo{person}{Feiyu Duan}, \bibinfo{person}{Liqun He}, \bibinfo{person}{Yutao Mou}, \bibinfo{person}{Wangchunshu Zhou}, \bibinfo{person}{Jiaheng Liu}, \bibinfo{person}{Wenge Rong}, \bibinfo{person}{Zekun~Moore Wang}, \bibinfo{person}{Jian Yang}, \bibinfo{person}{Ge Zhang}, \bibinfo{person}{Junran Peng}, \bibinfo{person}{Zhaoxiang Zhang}, \bibinfo{person}{Songyang Zhang}, {and} \bibinfo{person}{Kai Chen}.} \bibinfo{year}{2024}\natexlab{}.
\newblock \bibinfo{title}{HelloBench: Evaluating Long Text Generation Capabilities of Large Language Models}.
\newblock
\showeprint[arxiv]{2409.16191}~[cs.CL]
\urldef\tempurl%
\url{https://arxiv.org/abs/2409.16191}
\showURL{%
\tempurl}


\bibitem[Rohman(1965)]%
        {rohman1965pre}
\bibfield{author}{\bibinfo{person}{D~Gordon Rohman}.} \bibinfo{year}{1965}\natexlab{}.
\newblock \showarticletitle{Pre-writing: The stage of discovery in the writing process}.
\newblock \bibinfo{journal}{\emph{College Composition \& Communication}} \bibinfo{volume}{16}, \bibinfo{number}{2} (\bibinfo{year}{1965}), \bibinfo{pages}{106--112}.
\newblock


\bibitem[Schmidgall et~al\mbox{.}(2025)]%
        {schmidgall2025agent}
\bibfield{author}{\bibinfo{person}{Samuel Schmidgall}, \bibinfo{person}{Yusheng Su}, \bibinfo{person}{Ze Wang}, \bibinfo{person}{Ximeng Sun}, \bibinfo{person}{Jialian Wu}, \bibinfo{person}{Xiaodong Yu}, \bibinfo{person}{Jiang Liu}, \bibinfo{person}{Michael Moor}, \bibinfo{person}{Zicheng Liu}, {and} \bibinfo{person}{Emad Barsoum}.} \bibinfo{year}{2025}\natexlab{}.
\newblock \bibinfo{title}{Agent Laboratory: Using LLM Agents as Research Assistants}.
\newblock
\showeprint[arxiv]{2501.04227}~[cs.HC]
\urldef\tempurl%
\url{https://arxiv.org/abs/2501.04227}
\showURL{%
\tempurl}


\bibitem[Tang et~al\mbox{.}(2024)]%
        {tang2024llms}
\bibfield{author}{\bibinfo{person}{Xuemei Tang}, \bibinfo{person}{Xufeng Duan}, {and} \bibinfo{person}{Zhenguang~G Cai}.} \bibinfo{year}{2024}\natexlab{}.
\newblock \showarticletitle{Are LLMs Good Literature Review Writers? Evaluating the Literature Review Writing Ability of Large Language Models}.
\newblock \bibinfo{journal}{\emph{arXiv preprint arXiv:2412.13612}} (\bibinfo{year}{2024}).
\newblock


\bibitem[Team et~al\mbox{.}(2024)]%
        {team2024gemini}
\bibfield{author}{\bibinfo{person}{Gemini Team}, \bibinfo{person}{Petko Georgiev}, \bibinfo{person}{Ving~Ian Lei}, \bibinfo{person}{Ryan Burnell}, \bibinfo{person}{Libin Bai}, \bibinfo{person}{Anmol Gulati}, \bibinfo{person}{Garrett Tanzer}, \bibinfo{person}{Damien Vincent}, \bibinfo{person}{Zhufeng Pan}, \bibinfo{person}{Shibo Wang}, {et~al\mbox{.}}} \bibinfo{year}{2024}\natexlab{}.
\newblock \showarticletitle{Gemini 1.5: Unlocking multimodal understanding across millions of tokens of context}.
\newblock \bibinfo{journal}{\emph{arXiv preprint arXiv:2403.05530}} (\bibinfo{year}{2024}).
\newblock


\bibitem[Wei et~al\mbox{.}(2022)]%
        {wei2022chain}
\bibfield{author}{\bibinfo{person}{Jason Wei}, \bibinfo{person}{Xuezhi Wang}, \bibinfo{person}{Dale Schuurmans}, \bibinfo{person}{Maarten Bosma}, \bibinfo{person}{Brian Ichter}, \bibinfo{person}{Fei Xia}, \bibinfo{person}{Ed~H. Chi}, \bibinfo{person}{Quoc~V. Le}, {and} \bibinfo{person}{Denny Zhou}.} \bibinfo{year}{2022}\natexlab{}.
\newblock \showarticletitle{Chain-of-thought prompting elicits reasoning in large language models}. In \bibinfo{booktitle}{\emph{Proceedings of the 36th International Conference on Neural Information Processing Systems}} (New Orleans, LA, USA) \emph{(\bibinfo{series}{NIPS '22})}. \bibinfo{publisher}{Curran Associates Inc.}, \bibinfo{address}{Red Hook, NY, USA}, Article \bibinfo{articleno}{1800}, \bibinfo{numpages}{14}~pages.
\newblock
\showISBNx{9781713871088}


\bibitem[Wu et~al\mbox{.}(2023)]%
        {wu2023autogen}
\bibfield{author}{\bibinfo{person}{Qingyun Wu}, \bibinfo{person}{Gagan Bansal}, \bibinfo{person}{Jieyu Zhang}, \bibinfo{person}{Yiran Wu}, \bibinfo{person}{Beibin Li}, \bibinfo{person}{Erkang Zhu}, \bibinfo{person}{Li Jiang}, \bibinfo{person}{Xiaoyun Zhang}, \bibinfo{person}{Shaokun Zhang}, \bibinfo{person}{Jiale Liu}, \bibinfo{person}{Ahmed~Hassan Awadallah}, \bibinfo{person}{Ryen~W White}, \bibinfo{person}{Doug Burger}, {and} \bibinfo{person}{Chi Wang}.} \bibinfo{year}{2023}\natexlab{}.
\newblock \bibinfo{title}{AutoGen: Enabling Next-Gen LLM Applications via Multi-Agent Conversation}.
\newblock
\showeprint[arxiv]{2308.08155}~[cs.AI]
\urldef\tempurl%
\url{https://arxiv.org/abs/2308.08155}
\showURL{%
\tempurl}


\bibitem[Wu et~al\mbox{.}(2025)]%
        {wu2025longeval}
\bibfield{author}{\bibinfo{person}{Siwei Wu}, \bibinfo{person}{Yizhi Li}, \bibinfo{person}{Xingwei Qu}, \bibinfo{person}{Rishi Ravikumar}, \bibinfo{person}{Yucheng Li}, \bibinfo{person}{Tyler Loakman}, \bibinfo{person}{Shanghaoran Quan}, \bibinfo{person}{Xiaoyong Wei}, \bibinfo{person}{Riza Batista-Navarro}, {and} \bibinfo{person}{Chenghua Lin}.} \bibinfo{year}{2025}\natexlab{}.
\newblock \bibinfo{title}{LongEval: A Comprehensive Analysis of Long-Text Generation Through a Plan-based Paradigm}.
\newblock
\showeprint[arxiv]{2502.19103}~[cs.CL]
\urldef\tempurl%
\url{https://arxiv.org/abs/2502.19103}
\showURL{%
\tempurl}


\bibitem[Xu et~al\mbox{.}(2024)]%
        {10.5555/3692070.3694345}
\bibfield{author}{\bibinfo{person}{Haoran Xu}, \bibinfo{person}{Amr Sharaf}, \bibinfo{person}{Yunmo Chen}, \bibinfo{person}{Weiting Tan}, \bibinfo{person}{Lingfeng Shen}, \bibinfo{person}{Benjamin Van~Durme}, \bibinfo{person}{Kenton Murray}, {and} \bibinfo{person}{Young~Jin Kim}.} \bibinfo{year}{2024}\natexlab{}.
\newblock \showarticletitle{Contrastive preference optimization: pushing the boundaries of LLM performance in machine translation}. In \bibinfo{booktitle}{\emph{Proceedings of the 41st International Conference on Machine Learning}} (Vienna, Austria) \emph{(\bibinfo{series}{ICML'24})}. \bibinfo{publisher}{JMLR.org}, Article \bibinfo{articleno}{2275}, \bibinfo{numpages}{21}~pages.
\newblock


\bibitem[Yao et~al\mbox{.}(2023a)]%
        {yao2023tree}
\bibfield{author}{\bibinfo{person}{Shunyu Yao}, \bibinfo{person}{Dian Yu}, \bibinfo{person}{Jeffrey Zhao}, \bibinfo{person}{Izhak Shafran}, \bibinfo{person}{Thomas~L. Griffiths}, \bibinfo{person}{Yuan Cao}, {and} \bibinfo{person}{Karthik Narasimhan}.} \bibinfo{year}{2023}\natexlab{a}.
\newblock \showarticletitle{Tree of thoughts: deliberate problem solving with large language models}. In \bibinfo{booktitle}{\emph{Proceedings of the 37th International Conference on Neural Information Processing Systems}} (New Orleans, LA, USA) \emph{(\bibinfo{series}{NIPS '23})}. \bibinfo{publisher}{Curran Associates Inc.}, \bibinfo{address}{Red Hook, NY, USA}, Article \bibinfo{articleno}{517}, \bibinfo{numpages}{14}~pages.
\newblock


\bibitem[Yao et~al\mbox{.}(2023b)]%
        {yao2023react}
\bibfield{author}{\bibinfo{person}{Shunyu Yao}, \bibinfo{person}{Jeffrey Zhao}, \bibinfo{person}{Dian Yu}, \bibinfo{person}{Nan Du}, \bibinfo{person}{Izhak Shafran}, \bibinfo{person}{Karthik~R Narasimhan}, {and} \bibinfo{person}{Yuan Cao}.} \bibinfo{year}{2023}\natexlab{b}.
\newblock \showarticletitle{ReAct: Synergizing Reasoning and Acting in Language Models}. In \bibinfo{booktitle}{\emph{The Eleventh International Conference on Learning Representations}}.
\newblock
\urldef\tempurl%
\url{https://openreview.net/forum?id=WE_vluYUL-X}
\showURL{%
\tempurl}


\bibitem[Yi et~al\mbox{.}(2025)]%
        {yi2024survey}
\bibfield{author}{\bibinfo{person}{Zihao Yi}, \bibinfo{person}{Jiarui Ouyang}, \bibinfo{person}{Zhe Xu}, \bibinfo{person}{Yuwen Liu}, \bibinfo{person}{Tianhao Liao}, \bibinfo{person}{Haohao Luo}, {and} \bibinfo{person}{Ying Shen}.} \bibinfo{year}{2025}\natexlab{}.
\newblock \bibinfo{title}{A Survey on Recent Advances in LLM-Based Multi-turn Dialogue Systems}.
\newblock
\showeprint[arxiv]{2402.18013}~[cs.CL]
\urldef\tempurl%
\url{https://arxiv.org/abs/2402.18013}
\showURL{%
\tempurl}


\bibitem[Zhang et~al\mbox{.}(2024)]%
        {zhang2023sentiment}
\bibfield{author}{\bibinfo{person}{Wenxuan Zhang}, \bibinfo{person}{Yue Deng}, \bibinfo{person}{Bing Liu}, \bibinfo{person}{Sinno Pan}, {and} \bibinfo{person}{Lidong Bing}.} \bibinfo{year}{2024}\natexlab{}.
\newblock \showarticletitle{Sentiment Analysis in the Era of Large Language Models: A Reality Check}. In \bibinfo{booktitle}{\emph{Findings of the Association for Computational Linguistics: NAACL 2024}}, \bibfield{editor}{\bibinfo{person}{Kevin Duh}, \bibinfo{person}{Helena Gomez}, {and} \bibinfo{person}{Steven Bethard}} (Eds.). \bibinfo{publisher}{Association for Computational Linguistics}, \bibinfo{address}{Mexico City, Mexico}, \bibinfo{pages}{3881--3906}.
\newblock
\href{https://doi.org/10.18653/v1/2024.findings-naacl.246}{doi:\nolinkurl{10.18653/v1/2024.findings-naacl.246}}


\bibitem[Zhao et~al\mbox{.}(2023)]%
        {zhao2023more}
\bibfield{author}{\bibinfo{person}{Zoie Zhao}, \bibinfo{person}{Sophie Song}, \bibinfo{person}{Bridget Duah}, \bibinfo{person}{Jamie Macbeth}, \bibinfo{person}{Scott Carter}, \bibinfo{person}{Monica~P Van}, \bibinfo{person}{Nayeli~Suseth Bravo}, \bibinfo{person}{Matthew Klenk}, \bibinfo{person}{Kate Sick}, {and} \bibinfo{person}{Alexandre~LS Filipowicz}.} \bibinfo{year}{2023}\natexlab{}.
\newblock \showarticletitle{More human than human: LLM-generated narratives outperform human-LLM interleaved narratives}. In \bibinfo{booktitle}{\emph{Proceedings of the 15th Conference on Creativity and Cognition}}. \bibinfo{pages}{368--370}.
\newblock


\bibitem[Zheng et~al\mbox{.}(2023)]%
        {zheng2023judging}
\bibfield{author}{\bibinfo{person}{Lianmin Zheng}, \bibinfo{person}{Wei-Lin Chiang}, \bibinfo{person}{Ying Sheng}, \bibinfo{person}{Siyuan Zhuang}, \bibinfo{person}{Zhanghao Wu}, \bibinfo{person}{Yonghao Zhuang}, \bibinfo{person}{Zi Lin}, \bibinfo{person}{Zhuohan Li}, \bibinfo{person}{Dacheng Li}, \bibinfo{person}{Eric~P. Xing}, \bibinfo{person}{Hao Zhang}, \bibinfo{person}{Joseph~E. Gonzalez}, {and} \bibinfo{person}{Ion Stoica}.} \bibinfo{year}{2023}\natexlab{}.
\newblock \showarticletitle{Judging LLM-as-a-judge with MT-bench and Chatbot Arena}. In \bibinfo{booktitle}{\emph{Proceedings of the 37th International Conference on Neural Information Processing Systems}} (New Orleans, LA, USA) \emph{(\bibinfo{series}{NIPS '23})}. \bibinfo{publisher}{Curran Associates Inc.}, \bibinfo{address}{Red Hook, NY, USA}, Article \bibinfo{articleno}{2020}, \bibinfo{numpages}{29}~pages.
\newblock


\bibitem[Zheng et~al\mbox{.}(2025)]%
        {zheng2025deepresearcher}
\bibfield{author}{\bibinfo{person}{Yuxiang Zheng}, \bibinfo{person}{Dayuan Fu}, \bibinfo{person}{Xiangkun Hu}, \bibinfo{person}{Xiaojie Cai}, \bibinfo{person}{Lyumanshan Ye}, \bibinfo{person}{Pengrui Lu}, {and} \bibinfo{person}{Pengfei Liu}.} \bibinfo{year}{2025}\natexlab{}.
\newblock \bibinfo{title}{DeepResearcher: Scaling Deep Research via Reinforcement Learning in Real-world Environments}.
\newblock
\showeprint[arxiv]{2504.03160}~[cs.AI]
\urldef\tempurl%
\url{https://arxiv.org/abs/2504.03160}
\showURL{%
\tempurl}


\bibitem[Zhuang et~al\mbox{.}(2023)]%
        {zhuang2023toolqa}
\bibfield{author}{\bibinfo{person}{Yuchen Zhuang}, \bibinfo{person}{Yue Yu}, \bibinfo{person}{Kuan Wang}, \bibinfo{person}{Haotian Sun}, {and} \bibinfo{person}{Chao Zhang}.} \bibinfo{year}{2023}\natexlab{}.
\newblock \showarticletitle{ToolQA: a dataset for LLM question answering with external tools}. In \bibinfo{booktitle}{\emph{Proceedings of the 37th International Conference on Neural Information Processing Systems}} (New Orleans, LA, USA) \emph{(\bibinfo{series}{NIPS '23})}. \bibinfo{publisher}{Curran Associates Inc.}, \bibinfo{address}{Red Hook, NY, USA}, Article \bibinfo{articleno}{2180}, \bibinfo{numpages}{27}~pages.
\newblock


\bibitem[Zhuge et~al\mbox{.}(2024)]%
        {zhuge2024agent}
\bibfield{author}{\bibinfo{person}{Mingchen Zhuge}, \bibinfo{person}{Changsheng Zhao}, \bibinfo{person}{Dylan Ashley}, \bibinfo{person}{Wenyi Wang}, \bibinfo{person}{Dmitrii Khizbullin}, \bibinfo{person}{Yunyang Xiong}, \bibinfo{person}{Zechun Liu}, \bibinfo{person}{Ernie Chang}, \bibinfo{person}{Raghuraman Krishnamoorthi}, \bibinfo{person}{Yuandong Tian}, \bibinfo{person}{Yangyang Shi}, \bibinfo{person}{Vikas Chandra}, {and} \bibinfo{person}{Jürgen Schmidhuber}.} \bibinfo{year}{2024}\natexlab{}.
\newblock \bibinfo{title}{Agent-as-a-Judge: Evaluate Agents with Agents}.
\newblock
\showeprint[arxiv]{2410.10934}~[cs.AI]
\urldef\tempurl%
\url{https://arxiv.org/abs/2410.10934}
\showURL{%
\tempurl}


\end{thebibliography}

\appendix

\section{Question Examples}
Table~\ref{question_example} lists some examples, one example question for one type of finance research problem.

\begin{table}[H]
  \caption{Question Examples}
  \label{question_example}
  \scriptsize
  \begin{tabularx}{\columnwidth}{l X}
    \toprule
    Category & Example \\
    \midrule
    Stock-Specific Analysis & Deep research on Geely Automobile, especially the detailed assessment of the competitiveness of new vehicles and intelligent driving. \\
    \midrule
    Event Analysis & Please analyze the impact of the Trump administration's proposed new tariffs on A-share listed companies within the Apple supply chain. \\
    \midrule
    Stock Selection from Given Plate & Conduct an in-depth analysis of the eVTOL plate, requiring comprehensive conceptual explanation and fundamental analysis of the industry to which they belong. \\
    \midrule
    Sector Analysis & Analyze the 2025 development trends in the cosmetics industry, identifying potential high-impact breakout opportunities and examining the commercial/industrial dynamics driving them. \\
    \midrule
    Investment Morning Brief & Please compile a summary of today's investment morning brief, which should include key updates from yesterday till now, major industry events, overnight US stock market performance, and today's opening forecasts. \\
    \midrule
    Post-Market Recap & Analyze the fund flows of today's market, including the inflow and outflow of the institutional fund and the retail fund. \\
    \midrule
    Expert Opinion & Compile the latest opinions of the sell-side analysts over the past 12 hours and identify the three most frequently recommended stocks. \\
    \bottomrule
  \end{tabularx}
\end{table}

\section{Notations}
\label{app:notations}
Table~\ref{app:logic_tree_notation} lists the notations used in logic tree based metrics.

\begin{table}[H]
    \centering
    \caption{Notation for Logic Tree Based Metrics}
    \label{app:logic_tree_notation}
    \scriptsize
    \begin{tabularx}{\columnwidth}{l X}
        \toprule
        \textbf{Notation} & \textbf{Description} \\
        \midrule
        $S_{width}$ & The final score for Analysis Width. \\
        $\bar{N}_{child}$ & The average number of children for each node in the logic tree. \\
        \addlinespace 
        
        $S_{depth}$ & The final score for Analysis Depth. \\
        $S_{D_{max}}$ & The score component related to the maximum depth of the tree. \\
        $S_{\bar{D}_{leaf}}$ & The score component related to the average depth of leaf nodes. \\
        $D_{max}$ & The maximum depth of the logic tree. \\
        $\bar{D}_{leaf}$ & The average depth of all leaf nodes. \\
        \addlinespace
        
        $S_{info}$ & The final score for Information Density. \\
        $S_{nodes}$ & The score component related to the total number of nodes. \\
        $S_{density}$ & The score component related to the density of evidence. \\
        $N_{total}$ & The total number of nodes in the logic tree. \\
        $N_{evidence}$ & The number of nodes that contain evidence. \\
        $R_{evidence}$ & The ratio of evidence-containing nodes to the total number of nodes. \\
        \addlinespace
        
        $S_{rich}$ & The final score for Paragraph Richness. \\
        $w$ & The average number of words per subtitle in the report. \\
        \bottomrule
    \end{tabularx}
\end{table}

\section{Example Prompt}
This appendix provides illustrative examples of the prompt used to evaluate the reports.

\subsection{Logic Tree Extraction}
\label{appendix:logic_tree_extraction}
The prompt used to extract the logic tree.
\begin{lstlisting}[language=python, captionpos=t, caption={Prompt for Logic Tree Extraction}, label={lst:prompt_logic_extraction}]
"""
## BACKGROUND
You are an expert in logical analysis, specializing in identifying and constructing logical trees from text. To evaluate the quality of given answers based on metrics such as information volume, logical coherence, depth, and breadth, you must first identify the core arguments, sub-arguments, and evidence within the text and construct a logical tree to represent them.
## TASK
You will receive a query and the corresponding generated answer. This answer could be a casual reply, a research summary for a concept, or a professional research report. Your task is to carefully read the answer, identify all core arguments, sub-arguments, and evidence, and construct a logical tree to represent their relationships.
The core argument serves as the root node of the logical tree, sub-arguments are intermediate nodes, and evidence forms the leaf nodes. All root, intermediate, and leaf nodes must be extracted directly from the provided text. You must not add content not mentioned in the original text or make subjective inferences.
## DEFINITIONS
1.  **Evidence**: Objective facts presented in the text, including but not limited to statements of objective situations, events, or data.
2.  **Sub-argument**: A viewpoint or opinion derived from evidence or other sub-arguments through rigorous logical reasoning.
3.  **Core Argument**: A central claim derived from evidence and/or sub-arguments that condenses the essence of a significant portion, or the entirety, of the text. It represents the main takeaway of the report.
## PRINCIPLES FOR CONSTRUCTING THE LOGICAL TREE
1.  **Identify Evidence**: Read the text carefully to identify every piece of evidence. A single sentence may contain multiple pieces of evidence, or one piece of evidence may be only a part of a sentence. Strive for exhaustive identification without omission.
2.  **Construct Evidence Nodes**: When creating evidence nodes, you must **strictly adhere to the information provided in the original text**. Do not add extraneous information, make subjective inferences, or alter the original meaning.
3.  **Construct Intermediate Nodes (Sub-arguments)**:
    * A sub-argument can be formed in one of three ways:
        (i) Derived entirely from evidence (leaf nodes), representing a conclusion drawn directly from facts. These are typically found close to the leaves of the tree.
        (ii) Derived from a combination of evidence and other sub-arguments. These are common in the middle layers of the tree.
        (iii) Derived entirely from other sub-arguments, representing a higher-level conclusion. These are typically found closer to the root.
    * All deductions must be logically sound, with a clear and explicit relationship between a parent node and its children.
4.  **Construct the Root Node (Core Argument)**: The root node is primarily derived from sub-arguments, sometimes supplemented by key evidence.

User Query:
{query}
Article Content:
{report}

## OUTPUT FORMAT
Please strictly adhere to the following JSON format. Note:
1. All strings must be enclosed in English double quotes ("").
2. There should be no quotation marks within a string value.
3. Ensure the JSON syntax is correct, including commas, brackets, and braces.
4. Do not include any comments or additional explanations in the output.
5. The tree does not need to be strictly three levels deep; its depth and breadth are determined by the article's structure.
6. If the article contains multiple independent core arguments, the root node can be null, and the core arguments can be treated as first-level children.
7. Generally, the complexity of the logical tree should be proportional to the length and complexity of the answer. Keep this in mind during extraction.
## REQUIREMENTS
1.  Identify the core argument(s) of the text (root node).
2.  Identify the sub-arguments and evidence that support the core argument(s) (child nodes).
3.  Statements supported by data are classified as evidence; all others are considered arguments.
4.  Leaf nodes can be either arguments or evidence, depending on the specific context and the rule above.
"""
\end{lstlisting}

\end{document}